\documentclass[3p,12pt]{elsarticle}

\usepackage{amssymb}

\usepackage{hhline}
\usepackage{threeparttable}
\usepackage{multirow}
\usepackage{xcolor}
\usepackage{booktabs}
\usepackage{multicol}
\usepackage[ruled,vlined]{algorithm2e}
\usepackage[figuresright]{rotating} 
\usepackage{subfigure}
\usepackage{graphicx}
\usepackage{colortbl}
\usepackage{color}

\begin{document}

\begin{frontmatter}

\title{\textcolor[rgb]{0,0,0}{Mathematical Models and Reinforcement Learning based Evolutionary Algorithm Framework for Satellite Scheduling Problem} }

\author[]{Yanjie Song\fnref{fn1}}
\ead{songyj\_2017@163.com}

\fntext[fn1]{College of Systems Engineering, National University of Defense Technology, Changsha, Hunan, China, 410073}

\begin{abstract}
\textcolor[rgb]{0,0,0}{
For complex combinatorial optimization problems, models and algorithms are at the heart of the solution. The complexity of many types of satellite mission planning problems is NP-hard and places high demands on the solution. In this paper, two types of satellite scheduling problem models are introduced and a reinforcement learning based evolutionary algorithm framework based is proposed.
}
	
\end{abstract}

\end{frontmatter}

\section{Model for Electromagnetic Detection Satellite Scheduling Problem(EDDSP)}
\subsection{Problem Description}
The EDSSP problem is to designate a time-ordered task execution sequence for electromagnetic detection satellites \cite{song2023rl}. \textcolor[rgb]{0,0,0}{The goal is to maximize the detection sequence profit while satisfying various satellite constraints. For a satellite to successfully perform any mission, it needs to determine the on/off time of the receiver i.e. the start time and the end time of the mission. A series of parameter settings such as detection mode, frequency, bandwidth, and polarization mode must also be followed.}

The time range from the beginning to the end of the signal beam coverage of the electromagnetic satellite is called the visible time window. Since the electromagnetic satellite antenna can effectively detect a wide range of ground signals, to reduce noise and improve the detection accuracy, the angle between the signal source and the pointing direction of the satellite antenna needs to be within a certain range.

\textcolor[rgb]{0,0,0}{The detection quality is affected by two factors. On the one hand, the signal gain is affected by the angular relationship between the satellite antenna and the signal source. When the center of the satellite antenna passes directly above the signal source, the maximum signal gain can be obtained, and the signal gain is directly linked to the detection profit. Signal gain is inversely related to the angle between the antenna and the signal source. In other words, when the centerline of the signal source beam coincides with the extension line of the satellite antenna pointing direction, the detection effect is the best. On the other hand, the detection accuracy will also affect the detection profit. The detection accuracy is limited by the inherent capability of the receiver, and the number of frequency points will also have a direct impact on the detection accuracy. There is a positive correlation between the bandwidth and the number of frequency points, and the amount of data obtained by detection varies significantly depending on settings. The limited satellite storage capacity means that only a fraction of tasks can be set to the highest bandwidth, while many tasks need to be set to a smaller bandwidth.}

\subsection{Symbols and Variables}
$T$ : Set of tasks, a total of ${\left| T \right|}$ tasks. For task $task_j$, the following attributes are defined:

$es{t_j}$: The earliest available start time of the task.

$le{t_j}$: The latest available end time of the task.

$du{r_j}$: Duration of the task. 

$\theta _j^{\max }$: Maximum allowable detection angle of the task.

$degre{e_j}$ : Task importance level.

${m_j}$: The amount of data for the task.

${G_j}$: The signal gain can be obtained from the task.

$S$: Set of satellites, a total of ${\left| S \right|}$ satellites. For satellite ${s_i}$, the following attributes are defined:

$D$: Satellite antenna diameter.

$\eta $ : Antenna efficiency.

${O_i}$: Set of Orbits, a total of $\left| {{O_i}} \right|$ orbits belonging to satellite ${s_i}$.

$\beta $ : Satellite detection unit data volume.

$M$: Satellite storage capacity.

${\Gamma _{pol}}$ : Satellite polarization transition time.

${\Gamma _{mode}}$ : Satellite detection mode transition time.

${\Gamma _{band}}$: Satellite bandwidth setting transition time.

${\Gamma _{fre}}$: Satellite frequency band transition time.

$\Delta $ : Satellite load on/off time.

%
%
%

$TW$ : Set of time windows, with a total of $\left| {TW} \right|$ time windows. For the time window $t{w_{ijok}}$, the following attributes are defined:

$EV{T_{ijok}}$ : The earliest visible time of the task $j$ in the time window $k$ on orbit $o$ for satellite $i$.

$LV{T_{ijok}}$ : The latest visible time of the task $j$ in the time window $k$ on orbit $o$ for satellite $i$.

$\theta _{_{ijok}}^t$ : Detection angle  of the satellite $i$ at the time $t$ in the time window $k$ of the task $j$ on the orbit $o$.

$I$: A big integer.

Decision variables:

${x_{ijok}}$ : Whether the satellite $i$ is in the time window $k$ on the orbit $o$ whether the task $j$ is executed, if the task is executed, ${x_{ijok}} = 1$; otherwise, ${x_{ijok}} = 0$.

$s{t_{ijo}}$ : Start time of the satellite $i$ on the orbit $o$ to execute the task $j$.

\subsection{Mathematical Model}

\textbf{Assumptions:}
\begin{itemize}
\item All electromagnetic detection satellites have the same receivers and storage devices;

\item The detection process will not be affected by external factors;

\item The detection task is definite, and there will be no temporary changes or cancellations;

\item The satellite has sufficient energy during orbit;

\item Each task can be completed after one detection, without repeated detection.
\end{itemize}

The calculation formula of the detection profit that can be obtained by a single detection task is:

\begin{equation}
{G_j} = {G_0} \cdot {\left[ {\frac{{{J_1}\left( u \right)}}{{2u}} + 36\frac{{{J_3}\left( u \right)}}{{{u^3}}}} \right]^2}\left( {dBi} \right)\label{eq1}
\end{equation}

\begin{equation}
{G_0} = \eta \frac{{{\pi ^2}{D^2}}}{{{\lambda ^2}}}\left( {dBi} \right)\label{eq2}
\end{equation}

where $u = 2.07123\sin \left( \theta  \right)/\sin \left( {{\theta _{3dB}}} \right)$, ${{J_1}\left( u \right)}$ and ${{J_3}\left( u \right)}$ are the 1st and 3rd order Bessel functions of the first kind, respectively. $\theta$  is the angle between the satellite antenna and the center of the signal source, $\theta _{3dB}$ is the angle at which the antenna gain is attenuated by 3dB relative to the center of the beam, and the calculation formula as follows.
\begin{equation}{\theta _{3dB}} = 70\lambda /D\end{equation}

where $\lambda $ represents the wavelength, and $D$ indicates the diameter of the antenna.

In this paper, the bandwidth used by satellites to perform detection tasks is dynamically matched according to the priority of detection tasks. The importance of the task is high, and the bandwidth used is large so that the detection effect will be better. However, due to the limitation of satellite storage, the detection bandwidth needs to be scientifically set. The formula for setting the bandwidth according to the degree of importance ${degre{e_j}}$ is shown below.
\begin{equation}\varphi \left( {degre{e_j}} \right) = \left\{ {\begin{array}{*{20}{c}}
		{bandwidt{h_1}}&{degre{e_j} > 75}\\
		{bandwidt{h_2}}&{50 < degre{e_j} \le 75}\\
		{bandwidt{h_3}}&{25 < degre{e_j} \le 50}\\
		{bandwidt{h_4}}&{0 < degre{e_j} \le 25}
\end{array}} \right.\end{equation}

The bandwidth setting can also affect the detection profit, and the signal gain can be measured by the function $\Omega \left[ {\varphi \left( {degre{e_j}} \right)} \right]$. When the bandwidth of the detection task $j$ is set according to $\varphi \left( {degre{e_j}} \right)$, the amount of data generated per second is $\beta  \cdot \varphi \left( {degre{e_j}} \right)$. Combined with the task detection time $du{r_j}$, the total data amount of task $j$ can be obtained by:
\begin{equation}{m_j} = \beta  \cdot \varphi \left( {degre{e_j}} \right) \cdot du{r_j}\end{equation}

\textcolor[rgb]{0,0,0}{The main parameters of the electromagnetic detection task include frequency, bandwidth, polarization, and detection mode. The parameters of satellite $s_i$ for task $task_j$ are set as $\left(fre_{ij},band_{ij},pol_{ij},mode_{ij}\right)$. Where $fre_{ij}$ denotes the detection frequency, $band_{ij}$ denotes the bandwidth, $pol_{ij}$ denotes the polarization mode, and $mode_{ij}$ denotes the detection mode. The electromagnetic detection satellite needs to adjust the parameters of the onboard equipment when performing different tasks. The transition time between tasks is composed of four parts. The first part is the time required for the change of the polarization mode. The second part is the time required for the change of the detection mode. The third part is the time required for the change of the frequency, and the fourth part is the time required for the change of the bandwidth. In addition, it also takes a certain amount of time for each onboard equipment to be turned on and off, and this time interval $\delta$ must be satisfied between every two tasks. To simplify the constraint judgment, we introduce a new variable $tra{n_{ijj'}}$, which represents the total transition time. The transition time of the two tasks $j$ and ${j'}$ is as follows.}
\begin{equation}
\begin{array}{l}
	tra{n_{ijj'}} = \max \left\{ {{\Gamma _{fre}}\left( {fr{e_{ij}},fr{e_{ij'}}} \right),0,\Delta } \right.
	{\Gamma _{band}}\left( {ban{d_{ij}},ban{d_{ij'}}} \right),{\Gamma _{pol}}\left( {po{l_{ij}},po{l_{ij'}}} \right)\\
	\left. {{\Gamma _{mode}}\left( {mod{e_{ij}},mod{e_{ij'}}} \right)} \right\}
\end{array}
\end{equation}

where ${\Gamma _{pol}}$ is the satellite polarization transition time, 
${\Gamma _{mode}}$ is the satellite detection mode transition time, ${\Gamma _{band}}$ is the satellite bandwidth setting transition time, ${\Gamma _{fre}}$ is the satellite frequency transition time,  $\Delta $ is the satellite load on/off time.

\textcolor[rgb]{0,0,0}{The scheduling goal of the EDSSP problem is to obtain the highest detection profit.} The objective function is represented as follows.

\textbf{Objective function:}

\begin{equation}\max \sum\limits_{i \in S} {\sum\limits_{j \in T} {\sum\limits_{o \in {O_i}} {\sum\limits_{k \in TW} {{G_j}} } } }  \cdot \Omega \left[ {\varphi \left( {degre{e_j}} \right)} \right] \cdot {x_{ijok}}
\label{objective}
\end{equation}

where ${G_j}$ is signal gain can be obtained from the task, $\Omega \left[ {\varphi \left( {degre{e_j}} \right)} \right]$ is the gain due to the bandwidth setting, \textcolor[rgb]{0,0,0}{the product of ${G_j}$ and $\Omega \left[ {\varphi \left( {degre{e_j}} \right)} \right]$ represents the profit of the task.}

\textbf{Constraints:}

\begin{equation}s{t_{ijo}} \ge es{t_j} \cdot {x_{ijok}},\forall i \in S,j \in T,o \in {O_i},k \in TW\end{equation}
\begin{equation}\left( {s{t_{ijo}} + du{r_j}} \right) \cdot {x_{ijok}} \le le{t_j},\forall i \in S,j \in T,o \in {O_i},k \in TW\end{equation}
\begin{equation}\theta _{_{ijok}}^t \le \theta _j^{\max },\forall i \in S,j \in T,o \in {O_i},k \in TW,t \in \left[ {s{t_{ijo}},s{t_{ijo}} + du{r_j}} \right]
\label{angle}
\end{equation}
\begin{equation}s{t_{ijo}}  \ge EV{T_{ijok}} \cdot {x_{ijok}},\forall i \in S,j \in T,o \in {O_i},k \in TW\end{equation}
\begin{equation}\begin{array}{l}
	\left( {s{t_{ijo}} + du{r_j}} \right) \cdot {x_{ijok}} \le LV{T_{ijok}} ,\forall i \in S,j \in T,o \in {O_i},k \in TW
\end{array}
\label{tw time}
\end{equation}
\begin{equation}\begin{array}{*{20}{c}}
	{\sum\limits_{i \in S} {\sum\limits_{j \in T\backslash \left\{ {j'} \right\}} {\sum\limits_{k \in TW} {{m_j} \cdot {x_{ijok}}} } }  + {m_{j'}} \cdot {x_{ij'ok'}} \le M} {,\forall i \in S,j \in T,o \in {O_i},k \in TW}
\end{array}\end{equation}
\begin{equation}\begin{array}{*{20}{c}}
	{\left( {s{t_{ijo}} + du{r_j}} \right) \cdot {x_{ijok}} + tra{n_{ijj'}} \le s{t_{ij'o}} + I \cdot \left( {1 - {x_{ij'ok'}}} \right)} {,j \ne j',i \in S,j \in T,o \in {O_i},k \in TW}
\end{array}\end{equation}
\begin{equation}\sum\limits_{i \in S} {\sum\limits_{o \in {O_i}} {\sum\limits_{k \in TW} {{x_{ijok}}} } }  \le 1,\forall i \in S,o \in {O_i},k \in TW\end{equation}
\begin{equation}{x_{ijok}} \in \left\{ {0,1} \right\}\end{equation}

\textcolor[rgb]{0,0,0}{Constraints 1-2 indicate that the start time and end time of the task must be within the time range required. Constraint 3 indicates that the detection angle cannot exceed the task maximum angle requirement. Constraints 4-5 indicate that the start time and end time are within the visible time window. Constraint 6 indicates that the satellite cannot exceed the upper limit of the satellite storage capacity in each orbit. Constraint 7 indicates that the satellite must meet various transition time requirements to perform every two tasks. Constraint 8 indicates that each task can be executed at most once. Constraint 9 indicates the value range of the decision variable.}

\section{Model for Multi-satellite Joint Observation Planning Problem(MSJOPP) Considering Task Splitting}

Refer to \cite{song2022two}, this section describes the MSJOPP problem in text and mathematical model form . Firstly, the problem description and variables involved in the model are given.

\subsection{Symbols and Variables}
Symbols and variables in the multi-satellite joint observation planning model are described as follows.

${SAT}$: Set of satellites, with a total of ${\left| SAT\right|}$ satellites;

${\overline T}$: Set of satellite observation tasks, with a total of ${\left|\overline T\right|}$ tasks;

${{T}}$: Set of inseparable observation tasks, with a total of ${\left| T\right|}$ tasks;

${{\hat T }}$: Set of separable observation tasks, with a total of ${\left|\hat T\right|}$ tasks;

${{t_j}}$: An inseparable observation task;

${{\hat t_l}}$: A separable observation task, which can be split into ${{N_{{{\hat t}_l}}}}$ sub-tasks;

${TW}$: Set of time windows, with a total of ${\left| TW \right|}$ visible time windows;

${es{t_j}}$: Earliest allowable start time for observation task ${j}$;

${le{t_j}}$: Latest allowable end time for observation task ${j}$;

${du{r_j}}$: Required observation time for observation task ${j}$;

${{a_{i,j}}}$: Actual observation time of satellite ${i}$ for observation task ${j}$ ;

${{p_j}}$: Observation profit for task ${j}$ ;

${ev{t_{i,k}}}$: Earliest visible time of satellite ${i}$ for time window ${k}$;

${lv{t_{i,k}}}$: Latest visible time of satellite ${i}$ for time window ${k}$;

${{\hat a_{i,l,m}}}$: Sub-task  ${m}$ actual observation time of satellite ${i}$ for observation task ${l}$ ;

${\gamma }$: Minimum interval time between two observation tasks.

\textbf{Decision variables:}

The multi-satellite joint observation model needs to decide whether to execute inseparable observation tasks, separable observation tasks, and the specific execution time of each task/sub-task. Decision variables are described as follows.

${{x_{i,j,k}}}$: Observation task ${j}$ of satellite ${i}$  is arranged in time window ${k}$. When the task is successfully arranged ${{x_ {i,j,k}} = 1}$, otherwise, ${{x_{i,j,k}} = 0}$;

${{\hat x_{i,l,m,k}}}$: Sub-task ${m}$ of the separable observation task ${l}$ of satellite ${i}$ is arranged in time window ${k}$. When the sub-task is successfully arranged ${{\hat x_{i,l,m,k}}=1}$, otherwise, ${{\hat x_{i,l,m,k} }=0}$;

${{s_{i,j}}}$: Start time of satellite ${i}$ for observation task ${j}$;

${{\hat s_{i,l,m}}}$: Start time of satellite ${i}$ for separable observation task ${l}$'s ${m}$-th sub-task.

\subsection{Problem Description}

The multi-satellite joint observation planning problem needs to schedule a series of separable observation tasks and a series of inseparable observation tasks for multi-satellite resources. Before describing the problem in detail, the meaning of some terms involved in the problem is given first.

\textbf{Definition 1} (Separable observation task) Separable observation task means that when the required observation time of a task exceeds the length of each time window, this task can be split into multiple sub-tasks to be completed separately.

\textbf{Definition 2} (Inseparable observation task) Inseparable observation task means that an observation task must be completed by a single satellite in one observation process and cannot be divided into several sub-tasks.

\textbf{Definition 3} (Sub-task) Sub-task refers to a series of tasks obtained by splitting separable observation tasks through the task splitting method, and these sub-tasks together constitute the original observation task.

\textcolor[rgb]{0,0,0}{Here, a general description of the MSJOPP problem is given. Faced with a series of observation tasks, MSJOPP is to determine the best observation sequence. Determining the observation sequence requires the completion of the following two aspects, one is which satellite to complete each task, and the other is the specific time of the satellite to perform the task. For separable tasks, the algorithm also needs to determine the execution time of each subtask. Determining the execution time needs to meet a series of constraints, and whether the task can be observed is an important factor limiting the satellite's task execution. Since each satellite has its orbit, the payload on the satellite has a observation range, so the satellite can only observe a certain area near the orbit. The time range over which the satellite can observe the target area is called the satellite-to-earth visible time window. If the satellite wants to execute the observation task, it must be within one visible time window. If the task execution exceeds the visible time window, it cannot be completed. In addition to this constraint, other constraints are mainly related to the tasks submitted by users. These tasks have a series of time-related attributes. Attributes effectively limit the time range and duration of task execution. Tasks generally have timeliness requirements, which limit the earliest possible start and latest end times of the task, and the task execution needs to be within this time range. Meanwhile, the required observation time length also imposes restrictions on the time length of the task execution. No matter what kind of observation task it is, only when the actual observation time length is equal to the required observation time length can be counted as successful execution. Specifically, the split observation task requires that the accumulated value of the actual observation time length of multiple subtasks is equal to the required observation time length. A task can specify a start and end time when task requirements and time window constraints can be met. When a task is successfully observed, the profit can also obtain.}

In MSJOPP, an observation task can be represented by six tuple attribute ${\left[ {} es{t_j}, le{t_j}, du{r_j}, {a_{i,j}}, {p_j}, {s_{i,j}}\right]}$. The start time ${{s_{i,j}}, {\hat s_{i,l,m}}}$  of the separable/inseparable observation task ${{t_j}, {\hat t_l}}$ needs to be in the range of the earliest allowable start time of task ${es{t_j}}$ and the latest allowable end time ${le{t_j}}$. Correspondingly, the task start time ${{s_{i,j}}, {\hat s_{i,l,m}}}$ must also be in the range of the earliest visible time in one time window ${ev{t_{i,k}}}$ and the latest visible time ${lv{t_{i,k}}}$. In order to obtain observation profit ${{p_j}}$ of an inseparable task, the actual observation time ${{ a_{i,j}}}$ must equal to the required observation time ${du{r_j}}$. And accumulated sub-task observation time ${{\hat a_{i,l,m}}}$ of one separable observation task must equal to the original required observation time ${du{r_j}}$ .

Multi-satellite joint observation planning problem belongs to a category of satellite observation planning problems. The complexity of this problem is NP-hard \cite{berger2018graph}, so the optimal solution cannot be found in polynomial time. If an accurate algorithm is used for this type of planning problem, it is easy to fall into an exponential explosion. It is also difficult to find a feasible solution in a limited time. Numerous constraints and limited visual time windows increase the difficulty of solving the problem. Designing efficient and outstanding algorithms is the key to solving the problem.

In order to intuitively describe the multi-satellite joint observation planning problem, the following assumptions are made:

(1) Types of separable observation tasks and inseparable observation tasks have been specified in advance before planning and will not change during the planning process;

(2) Interval time between different tasks is a fixed value;

(3) Satellite in an abnormal working state is not considered due to satellite hardware equipment problems, and the task will not fail to execute;

(4) Each task can only be executed once at most, and the task will not be repeated periodically;

(5) There will be no temporary addition of new observation tasks or cancellation due to various factors in the planning process;

(6) Splitting will not add other additional task constraints for sub-tasks;

(7) Separable observation tasks and inseparable observation tasks do not have a difference in profits.

According to the description of this problem and the above assumptions, a mathematical model for multi-satellite joint observation planning can be constructed now. The mathematical model considers both the separable observation task and the inseparable observation task are given in the following part. Symbols and Variables in the model are given first.

\subsection{Mathematical Model}

In the multi-satellite joint observation model, there are two types of tasks, one is an inseparable observation task ${{T}}$ and the other is a separable observation task ${{\hat T}}$. Among them, the set of separable observation tasks ${{\hat T }}$ can obtain two task sets according to the task splitting algorithm, one is the sub-task set after splitting ${\hat T_1}$, and the other is the task set without splitting ${\hat T_2}$. The following relationships are satisfied between sets:

\begin{eqnarray}
	&\overline T = {T} \cup {\hat T}\\
	&{\hat T} =\hat  T_1 \cup \hat T_2
\end{eqnarray}

Refer to the research of \cite{chen2019mixed} , we construct a mixed-integer programming model (MIP) to describe the MOJSPP problem. The objective function of the multi-satellite joint observation planning problem is to maximize the profits of tasks. For this problem, there are two types of profits, including separable observation tasks and inseparable observation tasks. 
The completion of the inseparable observation task can be described by the scheduling of the original task. And for separable observation tasks, we need to consider the arrangement of each sub-task. Only when each sub-task is successfully scheduled the original task can be considered to be successfully executed. The objective functions of the multi-satellite joint observation planning problem are as follows:
\begin{eqnarray}
	\label{obj1}
	\max f = {f_1} + {f_2}
\end{eqnarray}
\begin{eqnarray}
	\label{obj2}
	{f_1} = \sum\limits_{i \in SAT} {\sum\limits_{j \in {T }} {\sum\limits_{k \in TW} {{x_{i,j,k} }}}} \cdot {p_j}
\end{eqnarray}
\begin{eqnarray}
	\label{obj3}
	{f_2} = \sum\limits_{i \in SAT} {\sum\limits_{l \in {\hat T }} {\sum\limits_{k \in TW} {\min \left( {{\hat x_{ i,l,1,k}},...,{\hat x_{i,l,m,k}}} \right)}}} \cdot {p_l}
\end{eqnarray}
\begin{eqnarray}
	{x_{i,j,k}},{\hat x_{i,l,m,k}} \in \left\{ {0,1} \right\}
\end{eqnarray}

Among them, ${f_1}$ represents the profit of inseparable observation task sequence, ${f_2}$ represents the profit of separable observation task sequence. ${{\min \left( {{\hat x_{i,l,1,k} },...,{\hat x_{i,l,m,k}}} \right)}}$ describes that original task . When it is is successfully planned, namely:
\begin{eqnarray}
	{\min \left( {{\hat x_{i,l,1,k}},...,{\hat x_{i,l,m ,k}}} \right)}=1
\end{eqnarray}

Factors that affect the execution of the observation task come from the requirement of tasks and the ability of satellites. Constraints of the multi-satellite joint observation planning model are as follows.

Constraints:

1. Actual start time of an observation task must be later than the earliest allowable start time:

\begin{eqnarray}
	es{t_j} \cdot {x_{i,j,k}} \le {s_{i,j}} ,{t_j} \in {T}
\end{eqnarray}

2. Actual end time of an observation task must be earlier than the latest allowable end time:

\begin{eqnarray}
	\left( {{s_{i,j}} + {a_{i,j}}} \right) \cdot {x_{i,j,k}} \le le{t_j},{ t_j} \in {T}
\end{eqnarray}

3. Actual start time of a sub-task must be later than the earliest allowable start time of the separable task:
\begin{eqnarray}
	es{t_l} \cdot {\hat x_{i,l,m,k}}\le {\hat s_{i,l,m}} ,{\hat t_l} \in {\hat T}
\end{eqnarray}

4. Actual end time of a sub-task must be earlier than the latest allowable execution time of the separable task:
\begin{eqnarray}
	\left( {{{\hat s}_{i,l,m}} + {{\hat a}_{i,l,m}}} \right) \cdot {\hat x_{i,l, m,k}} \le le{t_l},{\hat t_l} \in {\hat T}
\end{eqnarray}

5. Task must start within the visible time window of the observation target:
\begin{eqnarray}
	ev{t_{i,k}} \cdot {x_{i,j,k}} \le {s_{i,j}} ,{t_j} \in {T },t{w_k} \in TW
\end{eqnarray}

6. Task must complete within the visible time window of the observation target:
\begin{eqnarray}
	\left( {{s_{i,j}} + {a_{i,j}}} \right) \cdot {x_{i,j,k}} \le lv{t_{i,k}},{t_j} \in {T },t{w_j} \in TW
\end{eqnarray}

7. Sub-task of a separable observation task must start within the visible time window of the observation target:
\begin{eqnarray}
	ev{t_{i,k}} \cdot {\hat x_{i,l,m,k}} \le {\hat s_{i,l,m}} ,{\hat t_{i,l,m}} \in { \hat T },t{w_k} \in TW
\end{eqnarray}

8. Sub-tasks of a separable observation task must complete within the visible time window of the observation target:
\begin{eqnarray}
	\left( {{{\hat s}_{i,l,m}} + {{\hat a}_{i,l,m}}} \right) \cdot {\hat x_{i,l, m,k}} \le lv{t_{i,k}},{\hat t_l} \in {\hat T },t{w_k} \in TW
\end{eqnarray}

9. Actual observation time of a task must be equal to the required observation time:
\begin{eqnarray}
	{a_{i,j}} = du{r_j},{t_j} \in {T}
\end{eqnarray}

10. Total actual observation time of each sub-task of a separable observation task must be equal to the required observation time of the separable observation task:
\begin{eqnarray}
	\sum\limits_{i \in SAT} {\sum\limits_{m \in {{\hat t}_l}}} {{{\hat a}_{i,l,m}}} = du{r_l}, {\hat t_l} \in {\hat T}
\end{eqnarray}

11. Any two observation tasks must meet the minimum interval time requirements:
\begin{eqnarray}
	{s_{i,j + 1}}-\left( {{s_{i,j}} + {a_{i,j}}} \right) \ge \gamma ,{t_j}, {t_{j + 1}} \in {T}
\end{eqnarray}

12. Any two sub-tasks must meet the minimum interval time requirements:
\begin{eqnarray}
	{\hat s_{i,l,m + 1}}-\left( {{{\hat s}_{i,l,m}} + {{\hat a}_{i,l,m}} } \right) \ge \gamma ,{\hat t_l} \in {\hat T}
\end{eqnarray}

13. Any sub-task of an inseparable observation task and a separable observation task must meet the minimum interval time requirement:
\begin{eqnarray}
	\left| {{{\hat s}_{i,l,m}}-\left( {{s_{i,j}} + {a_{i,j}}} \right)} \right| \ge \gamma ,{t_j} \in {T },{\hat t_l} \in {\hat T}
\end{eqnarray}
\begin{eqnarray}
	\left| {{s_{i,j}} - \left( {{{\hat s}_{i,l,m}} + {{\hat a}_{i,l,m}}} \right)} \right| \ge \gamma ,{t_j} \in {T },{\hat t_l} \in {\hat T }
\end{eqnarray}
\begin{eqnarray}
	\left[ {{s_{i,j}} - \left( {{{\hat s}_{i,l,m}} + {{\hat a}_{i,l,m}}} \right)} \right] \times \left( {{s_{i,j}} + {a_{i,j}} - {{\hat s}_{i,l,m}}} \right) \ge 0,{t_j} \in {T },{\hat t_l} \in {\hat T }
\end{eqnarray}

14. Each observation task can execute at most once:
\begin{eqnarray}
	\sum\limits_{k \in TW} {{x_{i,j,k}}} \le {1_i},{t_j} \in {T}
\end{eqnarray}

15. Each sub-task of a separable observation task can execute at most once:
\begin{eqnarray}
	\sum\limits_{m \in {{\hat t}_l}} {\sum\limits_{k \in TW} {{{\hat x}_{i,l,m,k} }}} \le 1,{\hat t_l} \in {\hat T},t{w_k} \in TW
\end{eqnarray}

Constraints 1-4 describe that execution must meet the time requirements. Constraints 5-8 describe that execution must ensure that the target is visible. Tasks must start and complete within the visible time window. Constraints 9-10 describe the relationship between actual observation time with the required observation time. Constraints 11-13 describe the minimum time interval limit of tasks. Constraints 14-15 describe the limit on times of executions. 
The multi-satellite joint observation task planning problem is similar to the parallel machine scheduling problem and the VRPTW problem. Similarities between MSJOPP and parallel machine scheduling or VRPTW are that they both need to arrange and determine the start time of each task. These two problems are also deterministic problems, and the mixed-integer programming model can be used for model solving. The obvious difference between the problem studied in this paper and similar domains is that each satellite can only have time windows for some of the tasks, which increases the complexity obviously. This is similar to some specific processes that must be completed by specific machines, but the time limit makes the MSJOPP problem more complicated.

It is not difficult to see that numerous constraints impose strict restrictions on the planning of an observation task. It is a feasible solution to solve the separable observation task and the inseparable observation task planning problem separately in terms of the difficulty of finding a solution and optimization effect.

\section{Reinforcement Learning Based Evolutionary Algorithm Framework(RL-EA) }
\begin{algorithm}[htp]
	\label{RL-EA}
	\caption{RL-EA}
	\LinesNumbered
	\KwIn{ population $P $, population size ${N_{p}} $, Fitness $F $, Reward matrix $R $, learning rate $\alpha  $, discount factor $\gamma $, Q-table $Q $, $\varepsilon $, control parameter $T$, Operators $O$}
	\KwOut{${Solution}$}
	Initialize algorithm parameters\;
	Initialize the population\;
	Set $t$=1\;
	\While{termination criterion is not met}{

		${{A_t}}\leftarrow$Use Q-Learning to choose the appropriate action $\left(Q_{t},\varepsilon,T,O\right)$\;
		$individual \leftarrow$Select Individual $\left({P_{t-1}},{F_{t-1}}\right)$\;
		${P_t}\leftarrow $Population Evolution$\left({P_{t-1}},{A_{t}},individual\right)$\;
		${F_{t}} \leftarrow $Fitness Evaluation $\left({P_{t}}\right)$ \;
		${R_{t}}\leftarrow$Compute Reward$\left({F_{t}}, {F_{t-1}}\right)$\;
		${Q_{t+1}}\leftarrow$Compute Q-values$\left( {{S_{t}},{A_{t}}},{R_{t}},\alpha,\gamma \right)$\;
		$t\leftarrow t + 1$\;
}\end{algorithm}
Evolutionary algorithms simulate the evolution process of biological populations and find high-quality solutions to problems in the form of population evolution. Population search brings a strong global search capability to evolutionary algorithms, which ensures its performance in solving large-scale complex problems.

The obvious disadvantage of evolutionary algorithms is that the search results are extremely sensitive to the parameter configuration in the algorithm. Parameter-sensitive features make many evolutionary algorithms have certain dependencies on problems and scenarios. When problems or scenarios change, parameters need to be adjusted or reset. The parameter setting process is likely to be time-consuming. To enhance the versatility of evolutionary algorithms in different scenarios, the workload of setting algorithm parameters is reduced. We use the Q-learning method to drive the algorithm to update autonomously and let the evolutionary algorithm select an appropriate search strategy based on the search performance to drive the population evolution. Without loss of generality, we present a general evolutionary algorithm framework based on reinforcement learning. The algorithm framework can be flexibly adjusted according to different evolutionary algorithms in actual use. The pseudo-code of the evolutionary algorithm framework based on reinforcement learning is shown in Algorithm Table 1.

The evolutionary algorithm framework based on reinforcement learning selects the appropriate action according to the state in the iterative search process of the population, chooses one action to obtain a new generation of the population, evaluates the performance of the population and the income after adopting the action, and updates the state for a new round of population evolution. Population evolution and state update are continuous in the process of population search, and there exists frequent information exchange between them.

As a general framework, population evolution can be flexibly set according to the specific evolutionary algorithm used and the problems to be solved. Input parameters, operators, control parameters, and other strategies in the algorithm may also be different according to the specific algorithm used. For example, the genetic algorithm uses control parameters to limit crossover and mutation, while the ant colony algorithm uses control parameters to limit pheromone update.

\bibliography{mybib}{}
\bibliographystyle{elsarticle-num}

\end{document}